\documentclass[10pt,twocolumn,letterpaper]{article}

\usepackage{iccv}
\usepackage{times}
\usepackage{epsfig}
\usepackage{graphicx}
\usepackage{amsmath}
\usepackage{amssymb}
\usepackage{caption}
\PassOptionsToPackage{prologue,dvipsnames}{xcolor}

\usepackage[pagebackref=true,breaklinks=true,letterpaper=true,colorlinks,bookmarks=false]{hyperref}

\iccvfinalcopy 

\ificcvfinal\fi

\usepackage{color}
\usepackage{xcolor}
\usepackage{graphicx}
\usepackage{booktabs}
\usepackage{amsmath, amsthm, amsfonts, amssymb}
\usepackage{mathrsfs}
\usepackage{textcomp}
\usepackage{epstopdf}
\usepackage{multirow}
\usepackage{wrapfig}
\usepackage{subfig}
\usepackage{booktabs} 
\usepackage[ruled]{algorithm2e} 
\usepackage{algorithmicx}  
\usepackage{algpseudocode}
\usepackage{ragged2e}
\usepackage[normalem]{ulem}
\usepackage{cleveref}
\usepackage{bm}
\usepackage{ulem}
\usepackage{makecell}
\usepackage{diagbox}

\SetAlFnt{\small}
\SetAlCapFnt{\small}
\SetAlCapNameFnt{\small}
\SetAlCapHSkip{0pt}

\definecolor{green}{rgb}{0, 0.5, 0}
\definecolor{orange}{rgb}{0.6, 0.3, 0.1}
\definecolor{red}{rgb}{1.0, 0.0, 0.0}
\definecolor{teal}{rgb}{0.0, 0.4, 0.4}
\definecolor{purple}{rgb}{0.65,0,0.65}
\definecolor{saffron}{rgb}{0.95,0.75,0.2}
\definecolor{turquoise}{rgb}{0.0,0.5,0.5}
\definecolor{brown}{rgb}{0.5, 0.16, 0.16}
\definecolor{brickred}{rgb}{.6, .2 .1}
\definecolor{coral}{rgb}{1,0.45,0.33}
\definecolor{newcolor}{rgb}{.8,.349,.1}

\begin{document}

	\title{EmoSet: A Large-scale Visual Emotion Dataset with Rich Attributes}
	

\vspace{-10pt}
\author{Jingyuan~Yang{\textsuperscript{1}},
	Qirui~Huang{\textsuperscript{1}},
	Tingting~Ding{\textsuperscript{1}},
	Dani~Lischinski{\textsuperscript{2}},
	Daniel~Cohen-Or{\textsuperscript{3}},
	Hui~Huang{\textsuperscript{1}\thanks{Corresponding author}}\\
	\textsuperscript{1}Shenzhen University 
	\textsuperscript{2}The Hebrew University of Jerusalem 
	\textsuperscript{3}Tel Aviv University \\	
	{\tt\small \{jingyuanyang.jyy, qrhuang2021, dingt6616, danix3d, cohenor, hhzhiyan\}@gmail.com}
	\vspace{-25pt}
}

\twocolumn[{
	\renewcommand\twocolumn[1][]{#1}
	\maketitle
	\begin{center}
		\centering
		\includegraphics[width=\linewidth]{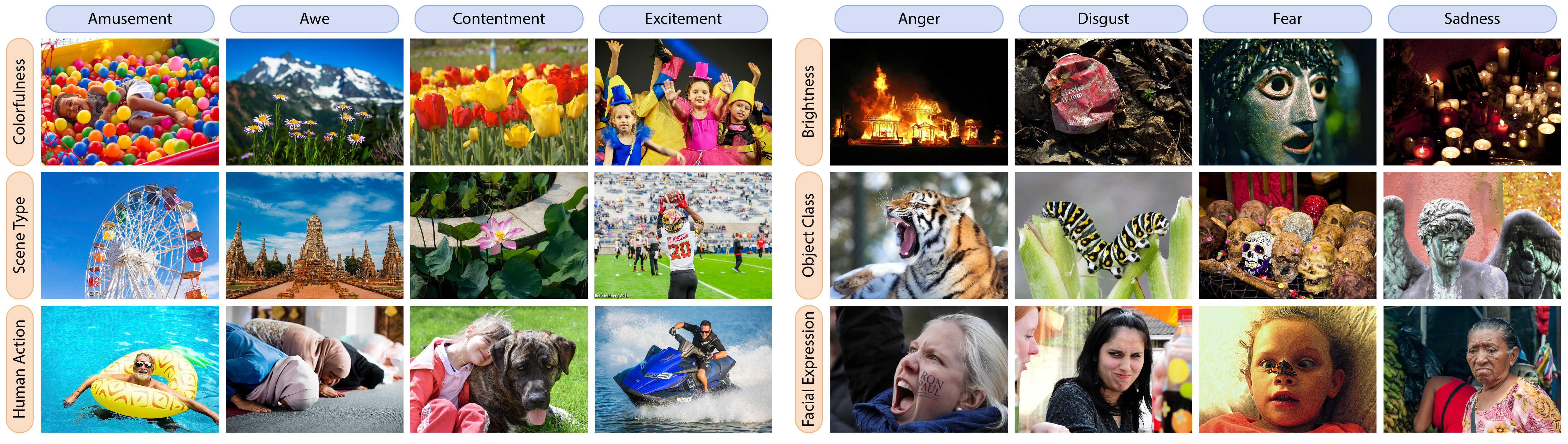}
		\captionof{figure}{EmoSet images are annotated with \textit{eight emotion categories} (\color[RGB]{84,127,231}{blue}\textcolor{black}{) and \textit{six emotion attributes} (}\color[RGB]{239,166,102}{orange}\textcolor{black}{), where different attributes may evoke different emotions.}}
		\label{fig:teaser}
	\end{center}
}]

\renewcommand{\thefootnote}{\fnsymbol{footnote}}

\ificcvfinal\thispagestyle{empty}\fi
\renewcommand{\thefootnote}{}

\begin{abstract}

	\vspace{-8pt}
	\footnotetext{*Corresponding author.}
	Visual Emotion Analysis (VEA) aims at predicting people's emotional responses to visual stimuli. This is a promising, yet challenging, task in affective computing, which has drawn increasing attention in recent years.
	Most of the existing work in this area focuses on feature design, while little attention has been paid to dataset construction.
	In this work, we introduce EmoSet, the first large-scale visual emotion dataset annotated with rich attributes, which is superior to existing datasets in four aspects: scale, annotation richness, diversity, and data balance.
	EmoSet comprises 3.3 million images in total, with 118,102 of these images carefully labeled by human annotators, making it five times larger than the largest existing dataset.
	EmoSet includes images from social networks, as well as artistic images, and it is well balanced between different emotion categories.
	Motivated by psychological studies, in addition to emotion category, each image is also annotated with a set of describable emotion attributes: brightness, colorfulness, scene type, object class, facial expression, and human action, which can help understand visual emotions in a precise and interpretable way.
	The relevance of these emotion attributes is validated by analyzing the correlations between them and visual emotion, as well as by designing an attribute module to help visual emotion recognition.
	We believe EmoSet will bring some key insights and encourage further research in visual emotion analysis and understanding.
	Project page: \url{https://vcc.tech/EmoSet}.
\end{abstract}

\section{Introduction}
\label{sec:intro}

\textit{Emotions are different ways to think that our mind uses to increase our intelligence}~\cite{minsky2007emotion}.
Much of the research in Artificial Intelligence (AI) has focused on designing human-like machines, while neglecting emotional intelligence. 
Since emotions are innate to human beings, AI systems should aim to better understand emotions, in order to succeed in mimicking human behavior. 
Affective computing~\cite{picard2000affective} is an emerging field that aims to identify, understand, and respond to human emotions. This field has seen significant progress in recent years, and has potential applications in areas such as education~\cite{yadegaridehkordi2019affective}, healthcare~\cite{yannakakis2018enhancing}, advertising~\cite{shukla2020recognition}, and safety~\cite{eyben2010emotion}.

Visual Emotion Analysis (VEA) is a promising, yet challenging, task in affective computing, aiming to predict emotional responses to visual stimuli.
For instance, when viewing the images in \Cref{fig:teaser}, one not only recognizes the visual elements therein, but also may experience emotional reactions. Furthermore, even though emotions are subjective, people tend to share similar reactions to the same external stimuli.
With the prevalence of social networks, users often choose to convey feelings via images shared on the internet.
Thus, VEA is an increasingly popular research topic within the computer vision field~\cite{yang2021solver, zhao2021affective}.
Advances in VEA may benefit high-level vision tasks (\eg, image aesthetic assessment~\cite{li2020personality}, stylized image captioning~\cite{guo2019mscap}, and image understanding~\cite{wu2019large}), as well as human-centered applications (\eg, opinion mining~\cite{li2019survey}, mental health~\cite{wieser2012reduced}, smart advertisement~\cite{sanchez2020opinion}, and hate detection~\cite{awal2021angrybert}).

Most of the work in VEA focused on feature design, covering hand-crafted features~\cite{machajdik2010affective,zhao2014exploring,borth2013large} and, more recently, learned ones~\cite{rao2016learning,yang2018weakly,yang2021solver}.
Based on art and psychological theories, hand-crafted features fail to cover all important factors in human emotions.
Although deep learning methods boost recognition performance significantly, the results are still unsatisfying.
In particular, while supervised deep learning methods often require large-scale labeled datasets, little attention has been paid to dataset construction.
Existing VEA datasets are usually unlabeled on a large scale or labeled on a relatively small scale~\cite{lang1999international, you2016building, peng2015mixed}.
Besides, only emotion labels are provided in most datasets.
Since emotions are abstract, a key problem is how to bridge the \textit{affective gap}~\cite{hanjalic2006extracting} between images and emotions with auxiliary information. 
We believe that a new and rich dataset is needed for further research and improvement in VEA.

To tackle the above issues, we introduce EmoSet, a large-scale visual emotion dataset, annotated with rich attributes.
EmoSet is superior to existing datasets in four aspects: scale, annotation richness, diversity, and data balance. 
The full (EmoSet-3.3M) dataset comprises 3.3 million machine retrieved and annotated images, among which there are 118,102 human-annotated ones (EmoSet-118K). The latter is five times larger than the widely-used FI dataset~\cite{you2016building}, as reported in \Cref{tab:dataset_statistic}.
Apart from emotions, our dataset is annotated with \emph{emotion attributes}.
Inspired by psychological studies~\cite{kurt2017modulation, brosch2010perception, ekman1993facial}, we propose a set of describable visual attributes to facilitate understanding \emph{why} an image evokes a certain emotion.
Considering the complexity of emotions, the attributes are designed to cover different levels of visual information, including brightness, colorfulness, scene type, object class, facial expression, and human action.
With these rich attribute annotations, we hope EmoSet will improve not only the recognition of visual emotions, but also their understanding.

By querying 810 emotion keywords based on Mikels model~\cite{mikels2005emotional}, we collect 3.3 million candidate images from four different sources to form the EmoSet-3.3M dataset.
A subset of EmoSet-3.3M is then labeled by human annotators, yielding the EmoSet-118K dataset. 
Compared with existing datasets, EmoSet contains diverse images covering both social and artistic types.
Furthermore, EmoSet-118K is well balanced between the eight emotion categories, each of which is represented with 10,660 to 19,828 images, as reported in \Cref{tab:emo_dist}.
We further analyze the correlations between our attributes and emotion categories, and demonstrate that some attributes are indeed strongly relevant to emotions.
In addition, to mine the emotion-related information from each attribute, we design an attribute module for visual emotion recognition, and validate it using several CNN backbones.

In summary, our contributions are:
\begin{itemize}
	\setlength{\itemsep}{0pt}
	\setlength{\parsep}{0pt}
	\setlength{\parskip}{0pt}
	
	\item EmoSet, the first large-scale visual emotion dataset with rich attributes, exceeding existing VEA datasets in terms of scale, annotation richness, diversity and data balance.
	
	\item A set of describable emotion attributes motivated by psychological studies, which help understand visual emotional stimuli in a precise and interpretable way. 
	
	\item A series of in-depth analyses on EmoSet, to explore how emotion attributes advance emotion understanding.
	Statistical analysis shows that correlations do occur between emotions and attributes, which is consistent with human cognition.
	
	\item An attribute module to facilitate visual emotion recognition.
	Experimental results and visualizations further validate the relevance of emotion attributes and show their potential in understanding visual emotions.
\end{itemize}

\section{Related Work}
\label{sec:rw}
%

\subsection{Visual Emotion Datasets}

In psychology, emotion models can be grouped into two types: Categorical Emotion States (CES), where emotions are described with discrete categories, and  Dimensional Emotion Space (DES), where a continuous space is used to represent emotions~\cite{zhao2016predicting}.
CES models are more popular in VEA for their simplicity and interpretability, including 2-category sentiment model, 6-category Ekman model~\cite{ekman1993facial}, and 8-category Mikels model~\cite{mikels2005emotional}.

Numerous datasets have been constructed to study people's different emotional reactions toward images~\cite{mikels2005emotional, peng2015mixed, you2016building}.
In \Cref{tab:dataset_statistic} we report various statistics of widely-used VEA datasets, as well as our new dataset, EmoSet.
Considering the diversity in image types, emotion models and dataset scales, below we discuss five of the existing datasets.

\textbf{IAPSa.}
IAPS~\cite{lang1999international} aims to investigate the possible relationships between emotions and visual stimuli.
IAPSa~\cite{mikels2005emotional} is a subset of IAPS, built on the Mikels model with eight emotion categories covering amusement, awe, contentment, excitement, anger, disgust, fear, and sadness.
IAPSa comprises 395 affective images, and is the first visual emotion dataset with discrete categories. 

\begin{table}[h] 
	\setlength{\abovecaptionskip}{0.1cm} 
	\setlength{\belowcaptionskip}{0.1cm}
	\tabcolsep=1cm
	\scriptsize
	\renewcommand\arraystretch{1.2}  
	\setlength\tabcolsep{2pt}   
	\centering
	\caption{Comparison between VEA datasets.}
	\label{tab:dataset_statistic}
	\begin{tabular}{lrcccc}
		\toprule[1pt]
		\multirow{2}{*}{Dataset}&\multirow{2}{*}{\#Image} & \multirow{2}{*}{\#Attribute} &\#Annotator & Model &Image\\
		&&&(per image)&(\#category)& type\\
		\midrule[0.5pt]
		IAPSa~\cite{mikels2005emotional} &  395     & - & -  & Mikels (8)  & Natural          \\
		Abstract~\cite{machajdik2010affective} & 280     & - & 14  & Mikels (8)      & Abstract         \\
		ArtPhoto~\cite{machajdik2010affective}  & 806    & - & -   & Mikels (8)     & Artistic         \\
		Twitter I~\cite{you2015robust}  & 1,269  & - & 5   & Sentiment (2)  & Social           \\
		Twitter II~\cite{borth2013large}   & 603    & - & 3  & Sentiment (2)   & Social           \\
		Emotion6~\cite{peng2015mixed}  & 1,980    & - & \textbf{15}  & Ekman (6) & Social           \\
		HECO~\cite{yang2022emotion} & 9,385 & - & - & - (8) & Social           \\
		Flickr~\cite{katsurai2016image}   & 60,745   & - & 3   & Sentiment (2)  & Social           \\
		Instagram~\cite{katsurai2016image} & 42,856 & - & 3   & Sentiment (2)  & Social           \\
		FI~\cite{you2016building}  & 23,308    & - & 5   & Mikels (8)     & Social           \\
		EMOTIC~\cite{kosti2017emotic}  & 18,316   & - & 3   & Ekman (6)      & Social           \\
		FlickrLDL~\cite{yang2017learning} & 10,700 & - & 11 & Mikels (8)  & Social           \\
		TwitterLDL~\cite{yang2017learning} & 10,045 & - & 8 & Mikels (8)  & Social           \\
		\textbf{EmoSet}     & \textbf{118,102}   & \textbf{6} & 10   & \textbf{Mikels (8)}  & \textbf{Social, Artistic}\\
		\bottomrule[1pt]
	\end{tabular}
\vspace{-10pt}
\end{table}

\textbf{ArtPhoto.}
There are a total of 806 artistic images in ArtPhoto~\cite{machajdik2010affective}, which is collected from an art sharing website by using emotion categories as the search keywords.

\textbf{Emotion6.}
Emotion6~\cite{peng2015mixed} has 1,980 images collected from Flickr.
Each image is labeled by 15 annotators based on Ekman model, covering six emotion categories, including happiness, anger, disgust, fear, sadness, and surprise. 

\textbf{Flickr and Instagram.}
Images in Flickr and Instagram~\cite{katsurai2016image} are crawled from the internet by searching emotional categories.
After labeling by crowd-sourced human annotation, 60,745 and 42,856 affective images are preserved with sentiment labels (\ie, positive or negative).
Specifically, each image is labeled by 3 workers, where the ground-truth is determined by a majority vote.

\textbf{FI.}
Another dataset based on Flickr and Instagram is FI~\cite{you2016building}. It is one of the largest scale visual emotion datasets to date, with a total of 23,308 labeled images.
Using eight emotion words in Mikels model as queries, FI collects candidate images from Flickr and Instagram.
Each collected image is then labeled by 5 Amazon Mechanical Turk (AMT) workers, and images having more than 3 votes are kept in the final dataset.
FI serves as one of the most widely-used datasets in VEA. 

In this work, we introduce EmoSet, which is five times larger than the FI dataset. In addition to the Mikels emotion category, each image is annotated with six comprehensively designed emotion attributes, including brightness, colorfulness, scene type, object class, facial expression and human action.
The images in our dataset come from more diverse sources and the dataset is more balanced across emotion categories, as reported in \Cref{tab:dataset_statistic,tab:emo_dist}.

\subsection{Visual Emotion Recognition}

Researchers have been engaged in VEA for two decades, with approaches ranging from the early traditional ones to the recent deep learning ones.
Machajdik~\etal~\cite{machajdik2010affective} extracts specific image features to predict emotions, \ie, color, texture, composition, and content inspired by psychology and art theory. 
Adjective Noun Pairs (ANPs) are introduced by Borth~\etal~\cite{borth2013large} to help learn visual emotions from a semantic level.
By extracting a set of principle-of-art-based emotional features like balance, emphasis, harmony, variety, gradation, and movement, Zhao~\etal~\cite{zhao2014exploring} proposes a method to deal with both classification and regression tasks.
Traditional methods fail to cover all important factors related to human emotions, leading to sub-optimal results.

Based on deep learning techniques, You~\etal~\cite{you2015robust} propose a progressive CNN (PCNN), and Rao~\etal~\cite{rao2016learning} build a multi-level deep representation network (MldrNet).
Early attempts usually focus on extracting holistic image features, while neglecting the importance of local regions.
Yang~\etal~\cite{yang2018visual} leverages object detection, as well as attention mechanism~\cite{yang2018weakly} to help emotion recognition.
With specially designed emotional features, Yang~\etal~\cite{yang2021stimuli} construct a network to learn emotions from different visual stimuli and to mine the correlations between them~\cite{yang2021solver}.
However, development in VEA is still unsatisfying due to low classification accuracy and the use of generic network design.
Considering the abstract nature of emotion, it is necessary to introduce auxiliary information to assist visual emotion recognition.
We believe that a large-scale dataset with rich annotations can help to mitigate the affective gap~\cite{hanjalic2006extracting} between images and emotion labels.

\section{Construction of EmoSet}
\label{sec:constructing}


\subsection{Data Collection}

To build EmoSet, we need to retrieve a large number of images from the Internet.
Since not all images are likely to arouse significant emotion, we construct a list of emotion keywords to help filter candidate images for our dataset.
Following previous work~\cite{machajdik2010affective, you2016building}, EmoSet is built on the widely-used Mikels model~\cite{mikels2005emotional} with eight categories, \ie, amusement, awe, contentment, excitement, anger, disgust, fear, sadness, where the former four are positive emotions and the latter four are negative ones.
Each of the eight emotion categories is first synonymized according to three widely-used English dictionaries: WordNet~\cite{miller1995wordnet}, Collins~\cite{hanks1979collins} and Roget's~\cite{roget1911roget}.
For instance, ``sadness'' is synonymized to words like ``depression, sorrow, mourn, despair, grieve''.
Since the number of retrievable images is often limited for each query, we combine the synonyms and further augment them with different parts of speech, aiming at retrieving
a large amount of data.
For example, ``amusement'' is augmented with other word forms like ``amuse, amuses, amused, amusing, amusingly''.
The final list contains 810 keywords, serving as queries to retrieve candidate images from the Internet.
For more details, please refer to the supplementary material.

In view of the fact that in most image-text pairs, the two modalities are in agreement, \ie, the textual tag or description indeed reflects the emotion that the image conveys, eqach retrieved image is automatically labeled with one of the eight emotions from which the query used to retrieve that image was derived.
For larger scale and richer diversity, EmoSet is collected from four different sources including openverse\footnotemark[1], pexels\footnotemark[2], pixabay\footnotemark[3] and rawpixels\footnotemark[4]. In total, 4.3 million images are collected, followed by voting in order to determine the labels of images with more than one emotion tag and removal of duplicates considering both file names and pixel-wise similarities, leaving 3.3 million images.
With such a large amount of images tagged with text descriptions, 
EmoSet-3.3M has great potential for weakly supervised learning~\cite{zhou2018brief}, vision-language modeling~\cite{radford2021learning} and multi-modal emotion analysis~\cite{tripathi2018multi}.

\footnotetext[1]{\url{https://wordpress.org/openverse/}}
\footnotetext[2]{\url{https://www.pexels.com/}}
\footnotetext[3]{\url{https://pixabay.com/}}
\footnotetext[4]{\url{https://www.rawpixel.com/}}

\subsection{Emotion Attributes Design}
\label{sec:attribute_design}

Aiming at figuring out some possible visual cues related to human emotions, we propose a set of describable emotion attributes inspired by psychological studies.
Since emotions are abstract and complex, emotion attributes are designed to cover different levels of visual information: low-level (\ie, brightness, colorfulness), mid-level (\ie, scene type, object class), and high-level (\ie, facial expression, human action).
We leverage both classic traditional methods and well-trained deep models to automatically predict attributes, which then constitute part of the automatic annotation, along with the emotion labels.

\begin{description}
	\setlength{\itemsep}{0pt}
	\setlength{\parsep}{0pt}
	\setlength{\parskip}{0pt}
	\item[Brightness:] The overall lighting level of an image has been proven to be essential in perceptual processing and is closely related to human emotions~\cite{kurt2017modulation}. We use discrete numerical values ranging from 0 (darkest) to 1 (brightest), with increments of 0.1, to quantify the brightness of an image~\cite{datta2006studying}. For more details, please refer to the supplementary material.
	\item[Colorfulness:]
	Psychological studies~\cite{ritchie2013perceived} suggest that there are correlations between the perceived colorfulness of an image and affect.
	Similarly to brightness, colorfulness is calculated, normalized and discretized to the range of 0 to 1~\cite{panetta2013no}.
	Specifically, 0 corresponds to black-and-white images, while 1 refers to the most colorful ones.
	\item[Scene type:]
	The scene depicted in an image is often considered as an important emotional stimulus~\cite{brosch2010perception}.
	We use a scene recognition model trained with Places365 \cite{zhou2017places}, a well-known benchmark for scene recognition.
	Out of 365 scene categories (\eg, sky, mountain, balcony, plaza, and church), we choose the top prediction as the scene type label for each image in our dataset.
	\item[Object class:] Psychologists have long investigated the relationships between objects and emotions~\cite{frijda2009emotion}. Thus, we associate object labels with each image in our dataset.
	Our object detection model is built on the OpenImagesV4 dataset \cite{kuznetsova2020open}.
	Considering that multiple objects may appear in an image and jointly evoke emotions, we associate with each image the three object classes predicted with the highest confidence. 
	\item[Facial expression:]
	Facial expression can undoubtedly influence visual emotion experience~\cite{ekman1993facial}, where people tend to empathize with the one in image.
	In Ekman model, there are six basic facial expressions: happy, angry, disgust, fear, sad, and surprise.
	We crop the largest face in the image and apply a model pre-trained on FER2013~\cite{goodfellow2013challenges} to obtain the facial expression label.
	\item[Human action:] 
	Some human actions stem from emotion and can also arouse emotion in an observer~\cite{zhu2002emotion}.
	Kinetics 400 is a large video dataset for human action recognition~\cite{kay2017kinetics}, which includes various actions like dining, water sliding, playing piano, barbecuing, and training a dog.
	Since Kinetics 400 is based on the video modality, we convert the image into a single-frame video as input and feed it into the UniformerV2 model~\cite{li2022uniformerv2} to predict the human action label.
\end{description}

\subsection{Human Annotation}

\begin{figure}
	\centering
	\includegraphics[width=\linewidth]{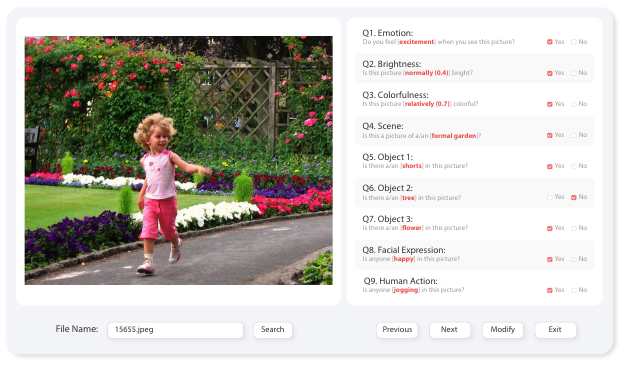}
	\caption{Graphical interface of annotation tool.
	}
	\label{fig:anno_tool}
	\vspace{-10pt}
\end{figure}

\begin{table*}
	\centering
	\scriptsize
	\caption{Comparison on image number per category, where {\color[RGB]{105,41,198}{purple}} indicates the maximum while {\color[RGB]{15,131,136}{green}} the minimum.}
	\label{tab:emo_dist}
	\renewcommand\arraystretch{1.08}
	\begin{tabular}{lccccccccl}
		\toprule
		Dataset & Amusement & Anger & Awe & Contentment & Disgust & Excitement & Fear & Sadness & Total\\
		\midrule 
		IAPSa & 37 (9\%) & \color[RGB]{15,131,136}{8 (2\%)} & 54 (14\%) & 63 (16\%) & \color[RGB]{105,41,198}{74 (19\%)} & 55 (14\%) & 42 (11\%) & 62 (16\%) & 395 \\
		Artphoto & 101 (13\%) & 77 (10\%) & 102 (13\%) & \color[RGB]{15,131,136}{70 (9\%)} & \color[RGB]{15,131,136}{70 (9\%)} & 105 (13\%) & 115 (14\%) & \color[RGB]{105,41,198}{166 (21\%)} & 806 \\
		Abstract & 25 (11\%) & \color[RGB]{15,131,136}{3 (1\%)} & 15 (7\%) & \color[RGB]{105,41,198}{63 (28\%)} & 18 (8\%) & 36 (16\%) & 36 (16\%) & 32 (14\%) & 228 \\
		FI & 4942 (21\%) & 1266 (5\%) & 3151 (14\%) & \color[RGB]{105,41,198}{5374 (23\%)} & 1658 (7\%) & 2963 (13\%) & \color[RGB]{15,131,136}{1032 (4\%)} & 2922 (13\%) & 23308 \\
		\textbf{EmoSet} & 19445 (16\%) & \textbf{\color[RGB]{15,131,136}{10660 (9\%)}} & 15037 (13\%) & 16337 (14\%) & 10666 (9\%) & \textbf{\color[RGB]{105,41,198}{19828 (17\%)}} & 13453 (11\%) & 12676 (11\%) & \textbf{118102} \\
		\bottomrule
	\end{tabular}	
	\vspace{-10pt}
\end{table*}


EmoSet-3.3M is automatically labeled by queries (\ie, emotions) and machines (\ie, attributes) without human participation.
To build a more carefully annotated dataset, we invite humans to help annotate and ask them to take the qualification tests first.
We ask the participants to take the empathy quotient test~\cite{baron2004empathy} to verify that they are sensitive to emotions, where annotators are qualified with a score greater than 30.
Subsequently, we randomly select 100 emotion-labeled images from the FI dataset to evaluate the classification accuracy of the participants, with a passing rate at 85\%.
We hired 60 annotators who passed all the above tests, thereby meeting our criteria.

There are three main challenges in visual emotion analysis: abstractness, ambiguity and subjectivity.
For abstractness, we introduce a set of attributes to help understand emotion in a more precise and interpretable way.
The annotation tool is presented in \Cref{fig:anno_tool}, where annotators are required to answer several questions on emotion (Q1) and attributes (Q2-Q9).
For example, annotators are asked ``Do you feel \textit{excitement} when you see this picture?'' (emotion) or ``Is this a picture of \textit{formal garden}?'' (attribute).
Since emotions are ambiguous, it is much easier for annotators to indicate whether an image evokes a specific emotion, rather than asking them to decide which emotion a given image evokes.
Fewer choices may lead to more accurate results.
Thus, we ask the annotators to verify both the emotion and the attribute labels for each image by answering ``yes'' or  ``no'', instead of selecting a specific category, following previous work~\cite{deng2009imagenet, you2016building}. 
To mitigate the subjectivity in emotion annotations, in EmoSet, each image is labeled by 10 annotators.
For each image, annotation results that reached a consensus of more than 7 out of 10 annotators are regarded as the final label.
In particular, images with more than 7 votes for ``yes'' in emotion label are preserved while others are deleted.
For more details, please refer to the supplementary material.
By the end, EmoSet-118K is carefully labeled with human annotations, where both emotion labels and attribute labels are provided.
Note that the analysis and evaluation in this paper are reported on EmoSet-118K.

\begin{figure}
	\centering
	\includegraphics[width=\linewidth]{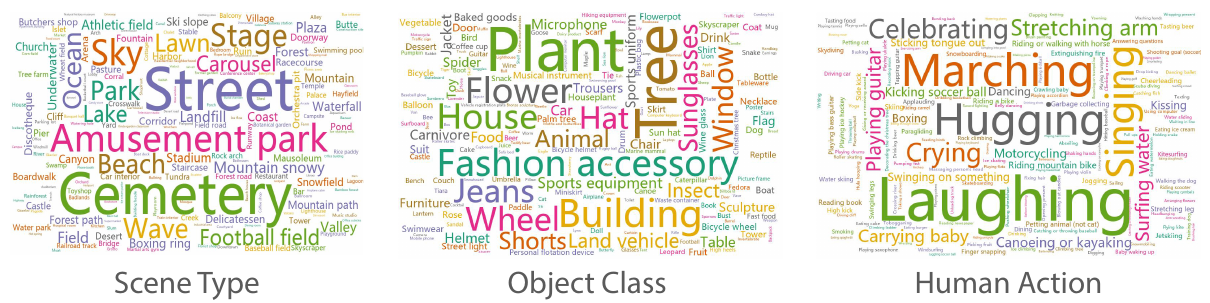}
	\vspace{-15pt}
	\caption{Word cloud distributions of scene type, object class and human action, where the larger the font, the higher the frequency it appears.	
	}
	\label{fig:att_cloud}
	\vspace{-15pt}
\end{figure}

\section{Analysis of EmoSet}
\label{sec:method}

\subsection{Properties of EmoSet}

EmoSet aims at constructing a comprehensive and interpretable dataset, which can help researchers to dive deep into visual emotions.
To our knowledge, this is the first large-scale VEA dataset that is also annotated with rich attributes, as shown in \Cref{tab:dataset_statistic}.
In general, EmoSet has advantages in four aspects compared with the existing datasets: scale, annotation richness, diversity, and data balance.

\begin{figure*}
	\centering
	\includegraphics[width=0.9\linewidth]{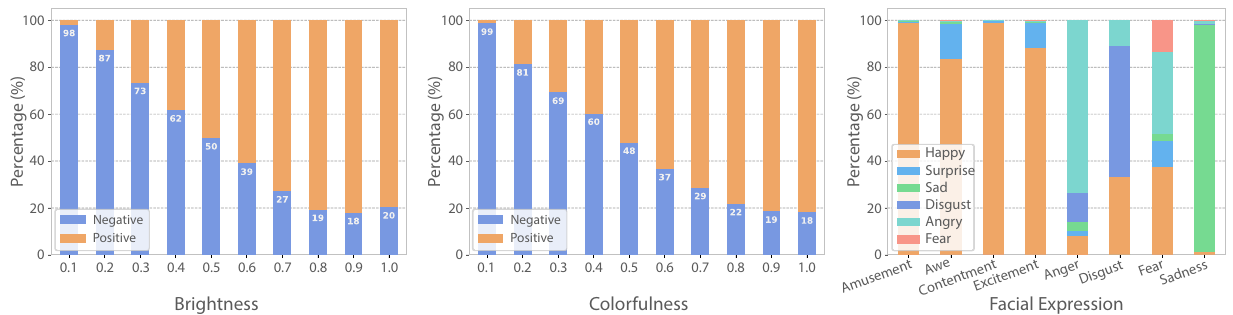}
	\vspace{-5pt}
	\caption{Histogram of brightness, colorfulness and facial expression, where different colors suggest different categories.}
	\label{fig:att_hist}
	\vspace{-5pt}
\end{figure*}

\begin{figure*}
	\centering
	\includegraphics[width=0.9\linewidth]{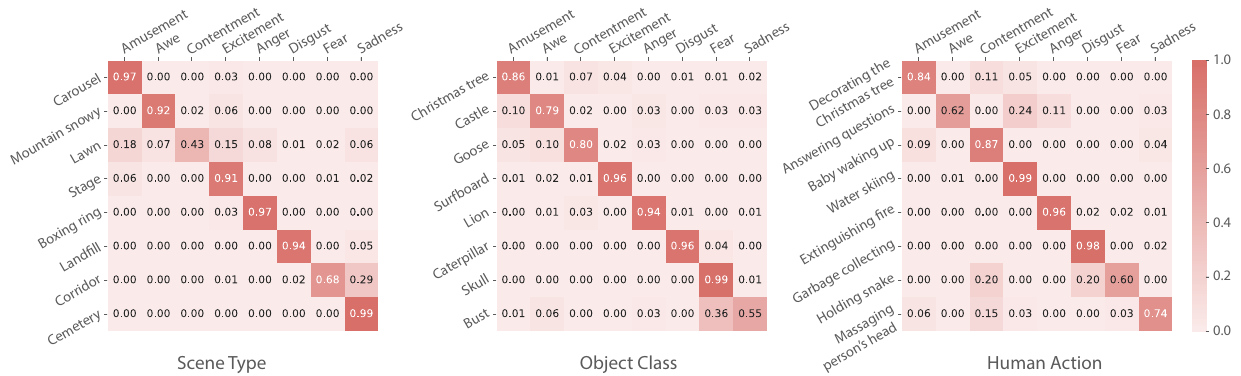}
	\vspace{-5pt}
	\caption{Correlation matrices of scene type, object class and human action, where numbers on the diagonal indicate the relationships between emotions and their top-1 attribute values.}
	\label{fig:att_cor}
	\vspace{-10pt}
\end{figure*}

\begin{description}
	\setlength{\itemsep}{0pt}
	\setlength{\parsep}{0pt}
	\setlength{\parskip}{0pt}
	\item[Scale:]
	Built with 3.3 million images with 118,102 human annotated ones, EmoSet is five times larger than the existing large-scale dataset FI (23,308).
	To our knowledge, it is the largest labeled visual emotion dataset in terms of the total number of images as well as the number of images per category.
	\item[Annotation richness:]	
	Based on Mikels model, EmoSet is labeled with eight emotion categories.
	In addition to emotion labels, we annotate EmoSet with six emotion attributes under different categories, which include \textit{brightness (10)}, \textit{colorfulness (10)}, \textit{scene type (365)}, \textit{object class (600)}, \textit{facial expression (6)}, and \textit{human action (400)}, to help understand emotions in a fine-grained manner.
	It is the first VEA dataset with attributes.
	In \Cref{fig:att_cloud}, we present the word cloud distribution of scene type, object class and human action.
	\item[Diversity:]
	Images are queried by 810 emotion keywords from four different sources, sharing a large data discrepancy.
	Different from the previous single-type emotion datasets, in EmoSet, there are images uploaded by social media users, as well as artistic work shared by professional photographers.
	For more details, please refer to the supplementary material.
	\item[Data balance:]
	As shown in \Cref{tab:emo_dist}, existing VEA datasets are unevenly distributed among different emotion categories, with the minimum and maximum numbers indicated by different colors. 
	In the Abstract dataset, for example, the \textit{anger} category is represented by only 1\% of the images, while 28\% represent \textit{contentment}.
	Data balance is essential for learning a good recognition model.
	Thus, our EmoSet is built with a balanced data distribution, where the number of images in each category is between 10,660 and 19,828.
\end{description}

\subsection{Emotion-Attribute Analysis}

Emotion attributes are designed to help visual emotion recognition as well as understanding.
To validate their effectiveness, we conduct several numerical experiments and visualizations on EmoSet to examine the relationships between attributes and emotions.

In Mikels model, amusement, awe, contentment, and excitement are considered \textit{positive} emotions, while anger, disgust, fear, and sadness are \textit{negative} ones.
The two polarities (\ie, positive, negative) can be seen as more basic emotional elements, than the eight more specific categories.
One might hypothesize that brighter and/or more colorful images are more likely to evoke a positive emotion. 
We verify this hypothesis in \Cref{fig:att_hist}, where we plot the breakdown of each brightness and colorfulness level into negative (blue) and positive (orange) emotions. Indeed, it may be seen that the proportion of images with a positive emotion label increases from left to right.


Our facial expression attribute is built upon Ekman model, where happy is positive, surprise is neutral, and other four (\ie, angry, disgust, fear, and sad) are negative.
In this experiment, we would like to see how facial expressions influence visual emotions.
In \Cref{fig:att_hist}, we show the breakdown of facial expressions for different visual emotions.
Unsurprisingly, all of the positive emotions exhibit a high correlation with a happy facial expression, while anger, disgust, and sadness are highly correlated with their corresponding facial expressions.
Interestingly, for fear, the top facial expression is happy, probably because the image contains a sinister or a spooky smile. 
The above experiment indicates that people are easily affected by the facial expressions present in the image, a manifestation of \textit{empathy}~\cite{elliott2011empathy}.

Each of the attributes scene type, object class, and human action, may have many different values, as in \Cref{fig:att_cloud}.
Obviously, some attribute values are strongly related to emotions (\eg, amusement park, cemetery, laughing, or crying), while others are not, such as sky, plant, tree, or window.
To discover emotion-related attribute values, we calculate the co-occurrence between each emotion-attribute pair and adopt the TF-IDF technique~\cite{salton1973construction}, where the importance of a value increases when it appears in a specific emotion and decreases with its appearance in the whole dataset.
\Cref{fig:att_cor} presents the correlation matrices between each emotion and its top-1 attribute value, where the large number on the diagonal suggests a strong relationship between them, with an average on 0.85 (scene type), 0.86 (object class) and 0.83 (human action).
The statistics in \Cref{fig:att_cor} are highly consistent with human cognition, indicating that some attribute values are indeed strongly related to emotions.
Once a certain attribute value appears, the image is much more likely to evoke the corresponding emotion.


\section{Evaluation of EmoSet}
\label{sec:results}

\subsection{Datasets Comparison}

\begin{table*}
	\centering
	\scriptsize
	\caption{Comparison on visual emotion recognition of EmoSet versus other VEA datasets with top-1 accuracy (\%).}
	\label{tab:exp_dataset}
	\renewcommand\arraystretch{1.18}
	\begin{tabular}{lcccccccc}
		\toprule
		Method & Twitter I-2 & Twitter II-2 & Flickr-2 & Instagram-2 & Emotion6-6 & FI-8 & EmoSet-2 & EmoSet-8\\
		\midrule
		AlexNet~\cite{krizhevsky2012imagenet} & 75.20 & 75.63 & 79.73 & 77.29 & 44.19 & 59.85 & \textbf{89.28} & \textbf{67.80} \\
		VGG-16~\cite{simonyan2014very} & 78.35 & 77.31 & 80.75 & 78.72 & 49.75 & 65.52 & \textbf{93.40} & \textbf{72.27}\\
		ResNet-50~\cite{he2016deep}  & 79.53 & 78.15 & 82.73 & 81.45 & 52.27 & 67.53 & \textbf{93.48} & \textbf{74.04}\\
		DenseNet-121~\cite{huang2017densely} & 80.71 & 78.99 & 84.87 & 83.76 & 53.79 & 67.24 & \textbf{92.92} &\textbf{72.32}\\
		\midrule
		WSCNet~\cite{yang2018weakly}  & 84.25 & 81.35 & 81.36 & 81.81 & 58.25 & 70.07 & \textbf{94.16} & \textbf{76.32}\\
		StyleNet~\cite{zhang2019exploring} & 81.50 & 80.67 & 85.02 & 84.53 & 59.60 & 68.85 & \textbf{93.93} & \textbf{77.11}\\
		PDANet~\cite{zhao2019pdanet} & 80.71 & 77.31 & 85.36 & 83.80 & 59.34 & 68.05 & \textbf{94.01} & \textbf{76.95}\\
		Stimuli-aware~\cite{yang2021stimuli} & 82.28 & 79.83 & 85.64 & 84.90 & 61.62 & 72.42 & \textbf{94.58} & \textbf{78.40}\\
		MDAN~\cite{xu2022mdan} & 80.24 & 83.05 & 84.26 & 83.52 & 61.66 & 76.41 & \textbf{93.71} & \textbf{75.75}\\
		\bottomrule
	\end{tabular}
\vspace{-10pt}
\end{table*}

\begin{table}
	\centering
	\scriptsize
	\caption{Ablation study of attribute module with different backbones on EmoSet, where results are reported in top-1 accuracy (\%).}
	\label{tab:exp_attribute}
	\renewcommand\arraystretch{1.18}
	\begin{tabular}{lcccccc}
		\toprule
		\multirow{2}{*}{Backbone} & \multicolumn{3}{c}{w/o pretrained} & \multicolumn{3}{c}{w/ pretrained} \\
		& w/o attr & w/ attr & $\Delta$ & w/o attr & w/ attr & $\Delta$ \\
		\midrule
		AlexNet & 46.44 & 55.80 & \textbf{$\uparrow$ 9.36} & 67.80 & 70.09 & \textbf{$\uparrow$ 2.29}\\
		VGG-16 & 48.51 & 56.51 & \textbf{$\uparrow$ 8.00} & 72.27 & 74.76 & \textbf{$\uparrow$ 2.49}\\
		ResNet-50 & 51.48 & 58.62 & \textbf{$\uparrow$ 7.14} & 74.04 & 76.60 & \textbf{$\uparrow$ 2.56}\\
		DensNet-121 & 53.09 & 60.77 & \textbf{$\uparrow$ 7.68} & 72.32 & 74.94 & \textbf{$\uparrow$ 2.65}\\
		\midrule
		Average & 49.88 & 57.93 & \textbf{$\uparrow$ 8.05} & 71.61 & 74.10 & \textbf{$\uparrow$ 2.50} \\
		\bottomrule
	\end{tabular}
	\vspace{-10pt}
\end{table}

Development in VEA has been greatly limited by the lack of large-scale, high-quality datasets, leading to unsatisfying results in visual emotion recognition, \ie, 59.60\% in Emotion6, 70.07\% in FI, as shown in \Cref{tab:exp_dataset}.
Consisting of 118,102 carefully annotated images, EmoSet aims at serving as an important dataset in VEA.  
To verify the quality of EmoSet, we compare it with other datasets, where results are reported in top-1 accuracy (\%).
We conduct experiments by leveraging both classic convolutional neural networks, \ie, AlexNet~\cite{krizhevsky2012imagenet}, VGG-16~\cite{simonyan2014very}, ResNet-50~\cite{he2016deep} and DenseNet-121~\cite{huang2017densely}, as well as VEA-oriented methods, \ie, WSCNet~\cite{yang2018weakly}, StyleNet~\cite{zhang2019exploring}, PDANet~\cite{zhao2019pdanet}, Stimuli-aware~\cite{yang2021stimuli} and MDAN~\cite{xu2022mdan}.
The classic convolutional neural networks are first pretrained on ImageNet~\cite{deng2009imagenet}, then fine-tuned and tested on each dataset respectively, while the VEA-oriented methods are trained and tested following their specific settings.
For EmoSet, we split the data into 80\% training set, 5\% validation set and 15\% test set, in accordance with that of FI.
Our network is trained by the adaptive optimizer Adam~\cite{kingma2014adam}, where the learning rate is set as the default one in each method.
Our experiments are implemented based on PyTorch~\cite{paszke2017automatic} and performed on an NVIDIA GTX 3090 GPU.
Notably, some datasets are labeled with two (\eg, Twitter I-2) or six (\eg, Emotion6-6) emotion categories, making it unfair to compare them with the eight-category EmoSet.
Therefore, we degenerate eight emotions to two sentiments, denoted as EmoSet-2, where amusement, awe, contentment and excitement are \textit{positive} and anger, disgust, fear and sadness are \textit{negative}.
In \Cref{tab:exp_dataset}, EmoSet reaches the best performance in both 2-sentiment and 8-category recognition tasks, compared with other VEA datasets.

\subsection{Attribute-aware Visual Emotion Recognition}

\begin{figure}
	\centering
	\includegraphics[width=\linewidth]{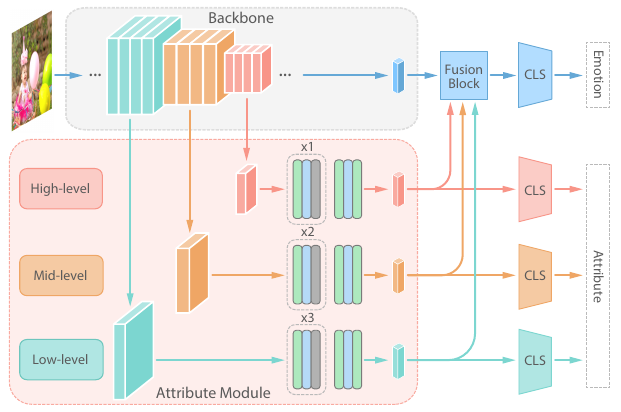}
	\caption{The proposed attribute module.}
	\label{fig:att_module}
	\vspace{-15pt}
\end{figure}

\begin{figure*}
	\centering
	\includegraphics[width=0.9\linewidth]{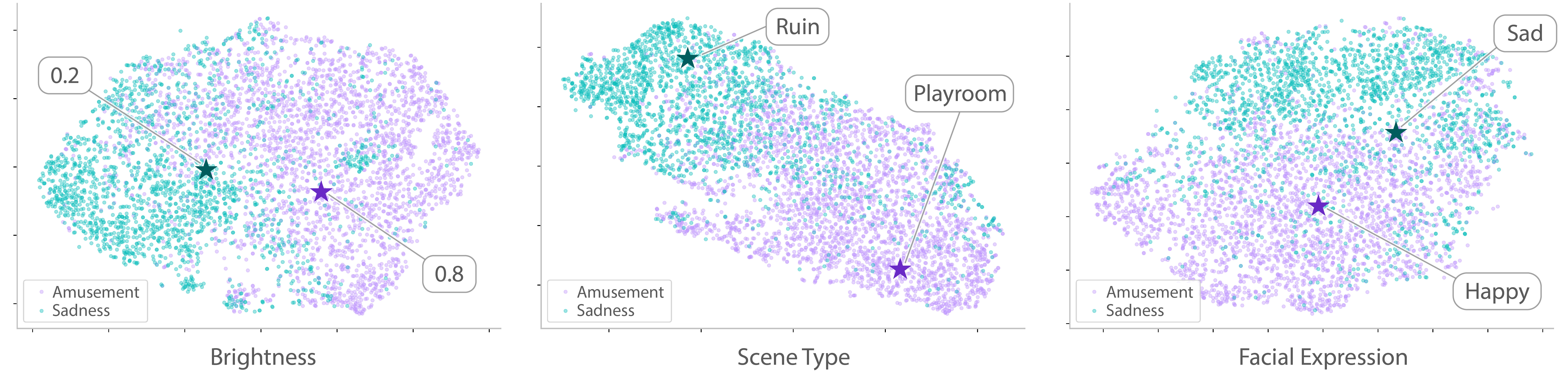}
	\vspace{-5pt}
	\caption{Scatter diagram of brightness, scene type and facial expression, where different colors represent attribute features of different emotions.}
	\label{fig:att_plot}
	\vspace{-15pt}
\end{figure*}

We propose an attribute module to facilitate visual emotion recognition, as shown in \Cref{fig:att_module}, which can be easily attached to a backbone network.
Attribute module is built with three branches, namely low-level, mid-level and high-level, to extract different visual information from a given image.
The extracted features are then sent to several lightweight convolutional layers, yielding attribute features.
Take backbone as main branch, features from other branches are fused with it to jointly predict visual emotions.
According to~\Cref{sec:attribute_design}, there are many options for selecting attributes.
In our experiments, we choose brightness, scene type and facial expression as representatives, which results are shown in \Cref{tab:exp_attribute}.
Each attribute branch is supervised by its ground-truth label, while the main branch is supervised by the emotion label.
The whole network is trained on EmoSet in an end-to-end manner.
Table~\ref{tab:exp_attribute} reports the top-1 accuracy (\%) under several conditions: with or without pretrained on ImageNet (\ie, w/ pretrained, w/o pretrained), with or without attribute module (\ie, w/ attr, w/o attr) and the differences between them (\ie, $\Delta$).
We conduct experiments with several widely-used backbones, including AlexNet, VGG-16, ResNet-50 and DenseNet-121.
The results indicate that emotion attribute boosts recognition performance to a large extent, especially when backbone has not been pretrained by ImageNet (\ie, 8.05\% on average).
The validity of emotion attributes has been verified in the above experiments, which serve as vital auxiliary information to help learn visual emotions.

\begin{table}
	\centering
	\scriptsize
	\setlength\tabcolsep{5.7pt}   
	\caption{Cross-dataset generalization between FI and EmoSet, where results are reported in top-1 accuracy (\%).}
	\label{tab:exp_generalize}
	\renewcommand\arraystretch{1.3}
	\begin{tabular}{l|cccccc}
		\specialrule{0.75pt}{1pt}{1.2pt}
		\multirow{2}{*}{\diagbox[dir=NW]{test}{train}} & \multicolumn{3}{c}{w/o pretrained} & \multicolumn{3}{c}{w/ pretrained} \\
		& FI & EmoSet & $\Delta$ & FI & EmoSet &  $\Delta$ \\
		\Xhline{0.5pt}
		
		FI & 40.62 & 35.26 & \textbf{$\downarrow$ 5.36} & 67.53 & 55.67 & \textbf{$\downarrow$ 11.86}\\
		EmoSet & 24.50 & 51.48 & \textbf{$\downarrow$ 26.98} & 53.95 & 74.04 & \textbf{$\downarrow$ 20.09}\\
		Artphoto & 18.23 & 25.06 & - & 29.28 & 32.26 & -\\
		\specialrule{0.75pt}{1.2pt}{1pt}
	\end{tabular}
\vspace{-10pt}
\end{table}

To verify how attributes assist emotion recognition, we further visualize the attribute features extracted from different branches, \ie, brightness, scene type and facial expression, as shown in~\Cref{fig:att_plot}.
Our visualizations are based on test set.
Each 2048-dimensional attribute feature is projected to a 2-dimensional vector by using t-Distributed Stochastic Neighbor Embedding (t-SNE)~\cite{van2008visualizing}, shown as a data point in the Cartesian coordinate. 
Taking amusement and sadness as examples, the scatter diagram is designated by two different colors, \ie, purple and green.
Obviously, data points can be separated by different emotions, especially in scene type, indicating that attribute features have learned some emotion-related information. 
We also visualize several emotion-related attribute values by calculating their cluster centers, which is denoted as stars in \Cref{fig:att_plot}.
In scene type, ``ruin'' and ``playroom'' are separated with a large distance, locating in sadness and amusement respectively.
Different attribute values fall in different emotional areas, which suggests that attribute features have been trained to distinguish both emotions and attributes jointly.
Visualization results in~\Cref{fig:att_plot} further proved the effectiveness of emotion attributes on assisting visual emotion recognition and understanding, which is also consistent with our human cognition.

\subsection{Cross-dataset Generalization}

To demonstrate the generalization ability of EmoSet, we conduct a cross-dataset validation between EmoSet and the large-scale FI in \Cref{tab:exp_generalize}.
We choose ResNet-50 as our backbone.
To validate the performance on a broader sense, we conduct two experiment settings: with or without pretrained on ImageNet, denoted as w/ pretrained, w/o pretrained. 
Trained on FI, the backbone meets a performance drop of 26.98\% (w/o pretrained) and 20.09\% (w/ pretrained), compared to the baseline of EmoSet.
Conversely, trained on EmoSet, the backbone meets a performance drop of 5.36\% and 11.86\%, correspondingly.
The above results illustrate that EmoSet is more capable to generalize to FI, compare with the opposite.
We further conduct validations by introducing a third-party test dataset, Artphoto, which consists of artistic images.
Since Artphoto is a small-scale dataset with 806 images, we only use it for test purpose.
In each setting, model trained on EmoSet performs better than that of FI, resulting from the diverse image types in EmoSet, \ie, social and artistic.    
Consisting of high-quality and diverse images, EmoSet is robust to generalize to other VEA datasets with a good visual emotion representation, which may bring new opportunities to visual emotion recognition.

\section{Conclusion}

	EmoSet is built with three main goals.
	The first one is to provide a large-scale, diverse and balanced VEA dataset, which may offer new opportunities for visual emotion recognition.
	Second, we believe that the rich annotated attributes can serve as auxiliary information to boost recognition performance.
	Most importantly, we hope that the comprehensively designed emotion attributes will encourage VEA researchers to turn their eyes from recognition to understanding, and to dive deep into visual emotion. 
	
	The underlying premise of our work is that each image evokes a single type of emotion. 
	In reality, different emotions may be evoked at the same time, considering the subjectivity of human emotions.
	Besides, our work is built upon Mikels emotion model with eight categories, following previous work.
	It is obvious that emotions are complex, and it is hard to precisely classify them into only a few discrete types.
	We will explore more in these directions.
	
	With 3.3 million images tagged with emotions and texts, EmoSet has the potential for weakly supervised learning, vision-language modeling and multi-modal emotion analysis.
	With rich attribute annotations, EmoSet also holds promise for visual emotion generation and editing.

\noindent\textbf{Acknowledgments:}
This work was supported in parts by NSFC (62161146005, U21B2023), DEGP Innovation Team (2022KCXTD025), Guangdong Science and Technology Program (2023A1515011440), Shenzhen Science and Technology Program (KQTD20210811090044003, RCJC20200714114435012), Israel Science Foundation (3441/21, 2492/20, 3611/21), and Guangdong Laboratory of Artificial Intelligence and Digital Economy (SZ).

{\small
\bibliographystyle{ieee_fullname}
\bibliography{EmoSet_ref}
}

\end{document}